\documentclass[runningheads]{llncs}
\usepackage{graphicx} 
\usepackage{amsmath} 
\usepackage{url}
\usepackage[T1]{fontenc}

\begin{document}
\title{A Turkish Educational Crossword Puzzle Generator\thanks{The funding for this paper was provided by the TAILOR project and the HumanE-AI-Net projects, both supported by the EU Horizon 2020 research and innovation program under GA No 952215 and No 952026, respectively.}}
%
%
\author{ Kamyar Zeinalipour\inst{1}\orcidID{0009-0006-3014-2511} \and Yusuf Gökberk Keptiğ\inst{1}\orcidID{0009-0002-2793-9061} \and Marco Maggini\inst{1}\orcidID{0000-0002-6428-1265} \and Leonardo Rigutini\inst{2}\orcidID{0000-0002-6309-2542} \and Marco Gori\inst{1}\orcidID{0000-0001-6337-5430}}

\authorrunning{Zeinalipour et al.}
%
\institute{University of Siena, Siena, Italy\\ \email{\{kamyar.zeinalipour2, marco.maggini, marco.gori\}@unisi.it
} \email{y.keptig@student.unisi.it}\and expert.ai\\ \email{lrigutini@expert.ai}\\}

\maketitle              
\begin{abstract}
This paper introduces the first Turkish crossword puzzle generator designed to leverage the capabilities of large language models (LLMs) for educational purposes. In this work, we introduced two specially created datasets: one with over 180,000 unique answer-clue pairs for generating relevant clues from the given answer, and another with over 35,000 samples containing text, answer, category, and clue data, aimed at producing clues for specific texts and keywords within certain categories. Beyond entertainment, this generator emerges as an interactive educational tool that enhances memory, vocabulary, and problem-solving skills. It's a notable step in AI-enhanced education, merging game-like engagement with learning for Turkish and setting new standards for interactive, intelligent learning tools in Turkish.
\end{abstract}
\section{Introduction}\label{sec:Introduction}

Crossword puzzles, designed with educational goals, blend puzzle-solving enjoyment with knowledge acquisition in fields like history, linguistics, and sciences, enhancing learners' vocabulary, spelling competence, and cognitive abilities such as memory retention and critical reasoning \cite{orawiwatnakul2013crossword,bella2023improving,mueller2018testing,zirawaga2017gaming,zamani2021use,dol2017gpbl}. They are particularly valuable in language learning, aiding in vocabulary building and the assimilation of technical jargon \cite{sandiuc2020use,yuriev2016crossword}. Recent advances in natural language processing (NLP), specifically LLMs, facilitate the creation of educational crosswords, offering high-quality clues and solutions based on user-suggested texts or keywords. This paper describes a novel application that leverages LLMs for generating Turkish educational crossword puzzles, supported by two new datasets: one with hand-crafted crossword puzzles and another with categorized Turkish texts, aimed at fostering the development and widespread use of educational crossword puzzles in Turkish learning environments. Through this work, the datasets and all of the developed 
models
\footnote{\url{https://huggingface.co/Kamyar-zeinalipour/llama7B_turkish_crossword_clue_gen}\\ \url{https://huggingface.co/Kamyar-zeinalipour/llama13B_turkish_crossword_clue_gen}\\ \url{https://github.com/KamyarZeinalipour/CW_Clue_Gen_tr}}
will be made available to the scientific community in open-source and free-to-download mode. The structure encompasses a literature review \ref{sec:relatedworks}, dataset description \ref{sec:dataset}, methodology \ref{sec:Methodology}, experiment analysis \ref{sec:Experiments}, and concluding remarks \ref{sec:conclusions}.

\section{Related Works}\label{sec:relatedworks}
The development of automatic crossword puzzle generation has evolved over time, incorporating various strategies from exploiting lexical databases and internet text analysis to adopting recent LLM-based techniques like fine-tuning and zero/few-shot learning. Initial attempts by Rigutini et al. leveraged advanced NLP for puzzle creation from online texts \cite{rigutini2008fully,rigutini2012automatic}, while other efforts focused on specific languages or themes, such as SEEKH for Indian languages \cite{arora2019automatic} and Esteche's work for Spanish-speaking audiences \cite{esteche2017automatic}. Recently,  \cite{zeinalipour2023arabicros,zeinalipour2023italian,zeinalipour2023building} shifted from manual crossword design to using pre-trained LLMs, generating puzzles in English, Arabic, and Italian, highlighting the power of computational linguistics in creating culturally diverse puzzles. In \cite{zugarini2024clue} they suggest a method for creating educational crossword clues dataset in English. However, the challenge of creating Turkish educational crossword puzzles was unaddressed. This study fills the gap by introducing a novel approach that employs advanced language models for generating engaging Turkish crossword puzzles, marking a significant advancement in educational tool development, especially for Turkish language learning.

\section{Dataset}\label{sec:dataset}
\begin{figure}[ht!]
    \centering
       \includegraphics[width=0.9\textwidth]{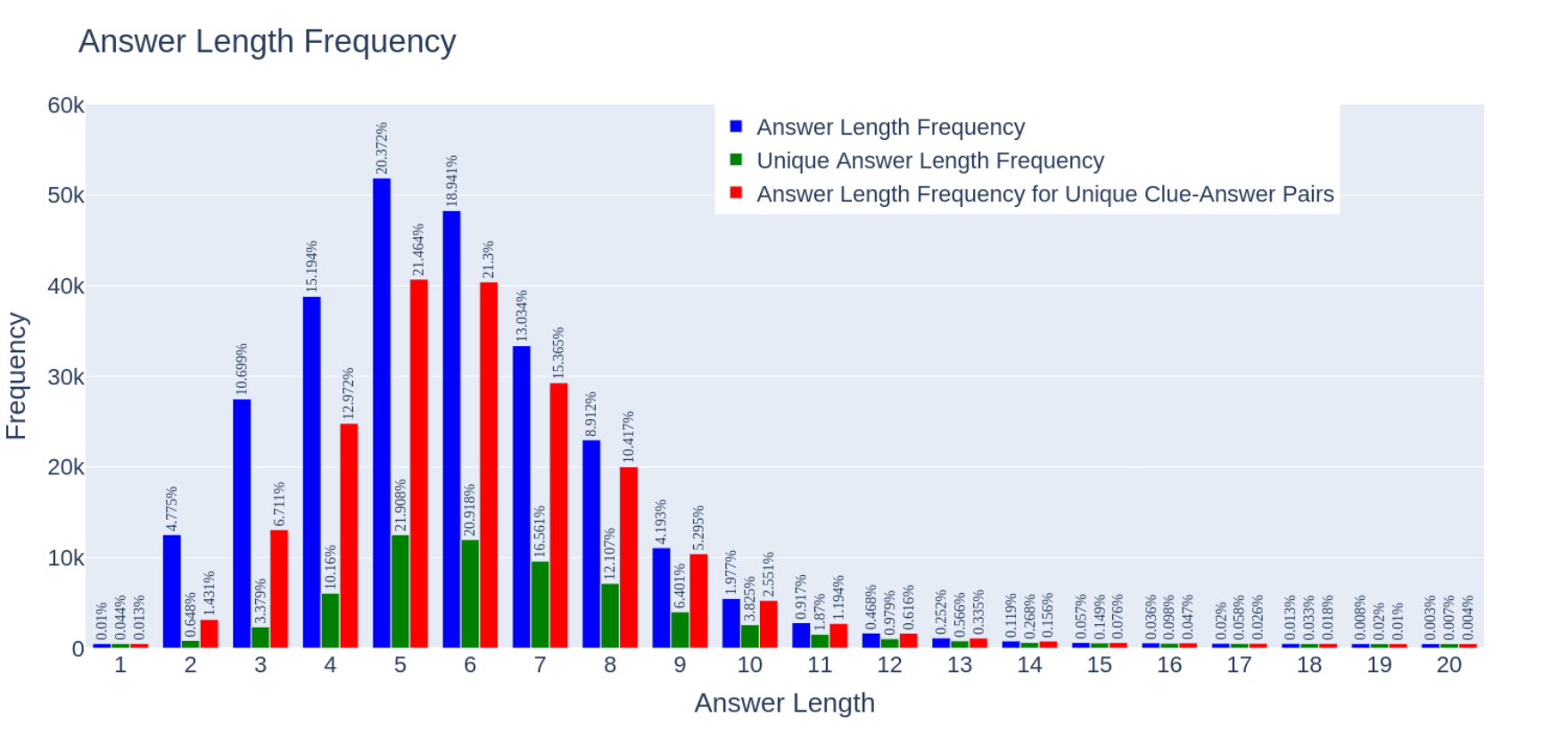}
    \caption{The dataset entries are showcased visually through the distribution of answer lengths. Blue bars represent all answer-clue pairs, green bars show the frequency of unique answers, and red bars display the frequency of unique answer-clue pairs}
    \label{fig:fig1}
\end{figure}

We constructed two datasets: the first consists of clue-answer pairs, and the second, a more elaborate compilation, merges text, answers, categories, and clues into one comprehensive dataset. The next section delves into the methods employed for their acquisition and compilation.
\paragraph{Turkish Answer-clue Pairs Dataset(TAC):}\footnote{\url{https://huggingface.co/datasets/Kamyar-zeinalipour/TAC}}
In this study, a comprehensive dataset of $252,576$ Turkish answer-clue pairs from online sources\footnote{\url{https://www.bulmaca-sozlugum.com}\\ \url{https://www.bulmacacozumleri.com}}, including newspapers like Habertürk, was compiled. These pairs, crafted by crossword experts, encapsulate $187,495$ unique entries, $54,984$ unique answers, and $178,431$ unique clues marking it a significant Turkish language resource. It's useful for applications like generating crossword clues for given answers.
In Figure \ref{fig:fig1}  we report a graphical representation of the distribution of answer's lengths. 

%

\begin{figure}[ht!]
    \centering
       \includegraphics[width=0.9\textwidth]{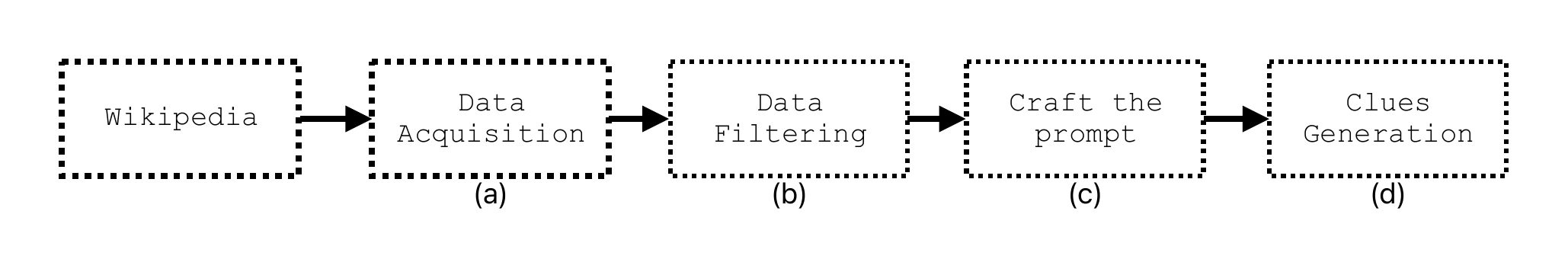}
    \caption{Diagram of the steps followed in the construction of the T4TAC dataset.}
    \label{fig:method_data}
\end{figure}

\begin{figure}[ht!]
    \centering    \includegraphics[width=0.9\textwidth]{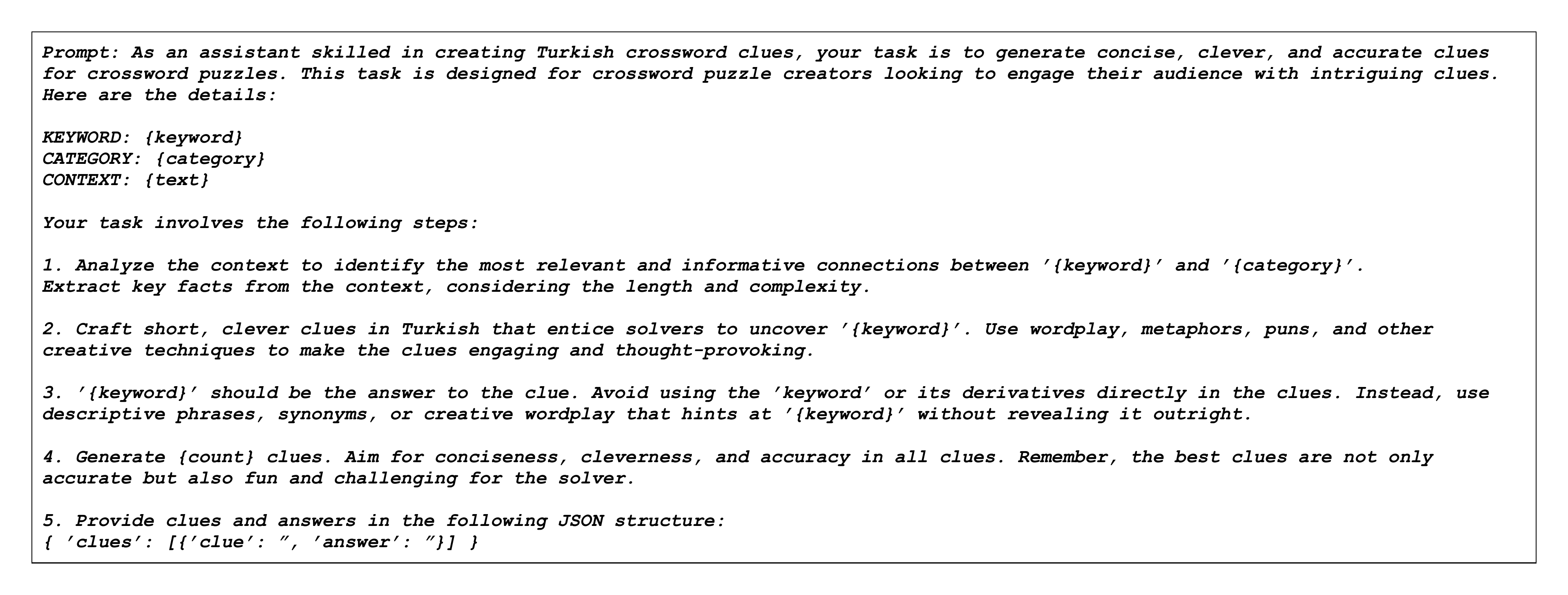}
   \caption{The prompt utilized in the study.}
    \label{fig:prompt}
\end{figure}

\begin{figure}[ht!]
    \centering
       \includegraphics[width=0.7\textwidth]{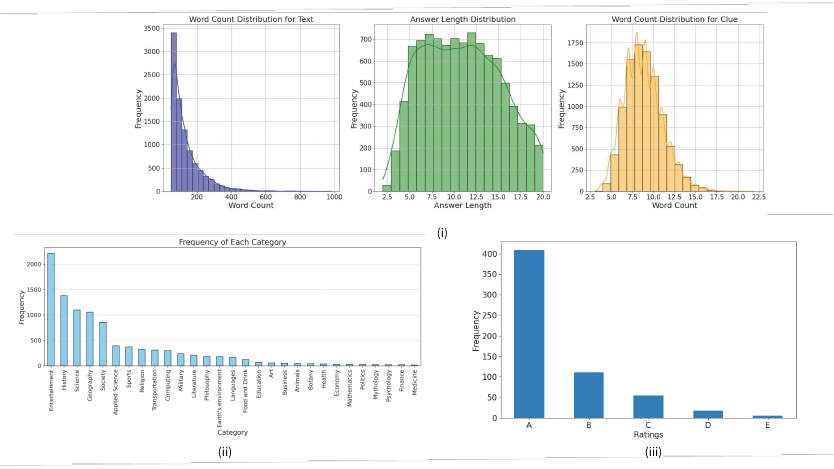}
    \caption{(i) Word and Character Length Distributions for Contexts, Outputs, and Keywords.(ii) Category Distributions of T4TAC. (iii) Human Evaluation for T4TAC.}
    \label{fig:dataset_distrubtions}
\end{figure}

\paragraph{Text for Turkish Answer-clue Pairs (T4TAC) Dataset:} \label{sec:data2} \footnote{\url{https://huggingface.co/datasets/Kamyar-zeinalipour/T4TAC}}
The second dataset in our study is crucial for developing models to autonomously produce crossword puzzle clues from the text and answer inputs in specific categories. This has significant potential in education, providing interactive teaching materials. The dataset creation process is outlined in Figure \ref{fig:method_data}. The workflow in Figure \ref{fig:method_data} is divided into stages important for the dataset development, starting with data collection where relevant text is gathered. This leads to data preprocessing to structure the text and keywords for training, followed by categorization to aid crossword clue generation. The Text for Turkish Answer-Clue pairs \textit{T4TAC} dataset is crucial for models that generate Turkish crosswords from text, benefiting educators. We outline the dataset construction phases.\\
\texttt{Data Acquisition and Filtering:} \label{sec:data_acquisition_filtering}
The extraction of Turkish Wikipedia articles targets the initial sections, emphasizing significant keywords and collecting metadata such as view counts, relevance scores, and URLs\footnote{\url{https://en.wikipedia.org/wiki/Wikipedia:Lists_of_popular_pages_by_WikiProject}}. The focus remains on keyword-rich introductory paragraphs. Quality assurance involves filtering pages by popularity, relevance, content length, and keyword specificity, excluding data unsuitable for crossword clues such as keywords outside 3 to 20 characters or containing special characters and numbers.\\
\texttt{Prompt Development and Educational Clue Creation:} \label{sec:prompt_education_clue_creation}
 The targeted prompt is crafted to generate informative and engaging Turkish crossword clues, illustrated in Figure~\ref{fig:prompt}. Utilizing the \textsc{self-instruct} framework~\cite{wang2022self} and \textit{GPT4-Turbo}, our method integrates contextual information with content, keywords, and categories for generating tailored Turkish educational clues. This approach results in clues that are educationally valuable and contextually relevant.\\
\texttt{Turkish Answer-Clue Dataset Compilation:} \label{sec:data_compilation}
From an initial set of $180,000$ Wikipedia pages, $9,855$ pages across $29$ themes were selected after filtering, producing over $35,000$ clues using \textit{GPT4-Turbo}. The dataset's content length ranges from 50 to 982 words, with clues generally spanning 5 to 15 words. Figure \ref{fig:dataset_distrubtions} (i) shows the distribution of word counts for both texts and clues and the lengths of answers. The dataset encompasses a broad spectrum of topics including dominant themes like "Entertainment," "History," and "Science," as illustrated in Figure \ref{fig:dataset_distrubtions} (ii).


\section{The Crossword Generation System} \label{sec:Methodology}
\begin{figure}[ht!]
    \centering
       \includegraphics[width=0.9\textwidth]{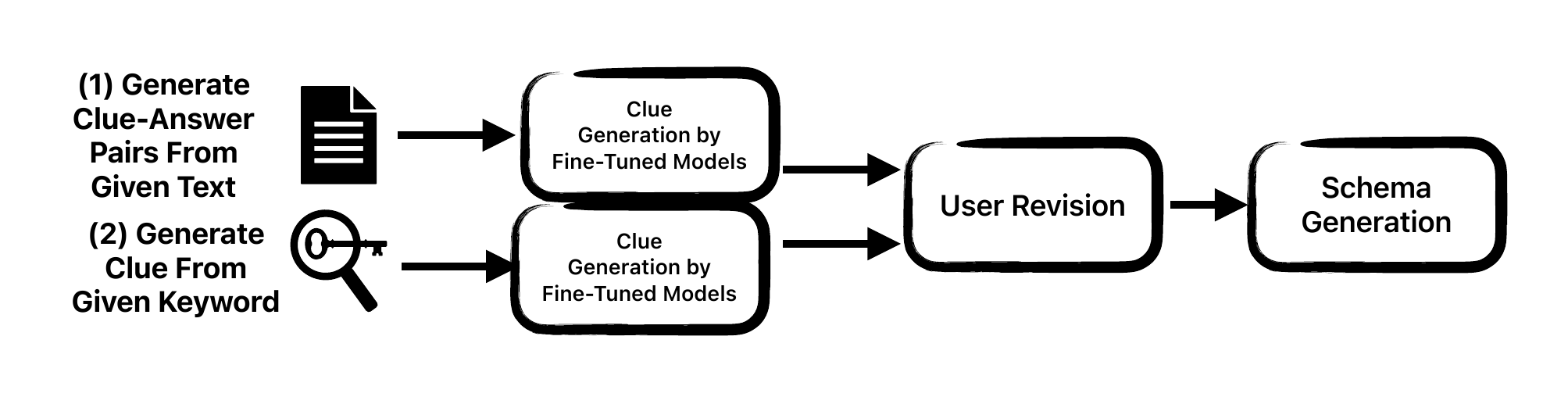}\\
    \caption{The scheme  of the crossword generation system}
    \label{fig:method}
\end{figure}

We developed a system for generating Turkish crossword puzzles from specified answers and texts, useful for educators. There are two main use cases: one where educators already have a list of keywords for creating puzzles without additional text; and another where they wish to generate crosswords from input texts by extracting keywords and clues automatically. We fine-tuned LLMs like \textit{GPT3.5-Turbo} \cite{brown2020language}, 
\textit{Llama-2-7b-chat-hf} and \textit{Llama-2-13b-chat-hf} \cite{touvron2023llama} using a specific dataset for clue generation, allowing educators to choose the best options. A specialized algorithm, detailed later, is used to create the crossword layout. The methodology is outlined in Figure \ref{fig:method}.

\paragraph{Fine-tuning LLMs to Generate Clues:}
Investigating how to generate crossword clues from answers and text, we improved language models, closely following the dataset described in Section \ref{sec:dataset}. For two distinct taks: creating crossword clues from provided answers using the \textit{TAC} dataset, and generating crossword clues from given text corresponding to specific categories using the \textit{T4TAC} dataset. We evaluated models like \textit{GPT3.5-Turbo} \cite{brown2020language}, \textit{Llama-2-7b-chat-hf} and \textit{Llama-2-13b-chat-hf} \cite{touvron2023llama}, using a mix of open-source and proprietary versions, including \textit{GPT3.5-Turbo}, after fine-tuning, to assess clue quality.
\paragraph{Schema Generator:} \label{sec:schema}
The crossword puzzle creation algorithm operates with input criteria like an answer list, puzzle size, and stopping rules to efficiently assemble puzzles. It starts by placing an initial word and adds others strategically, adjusting through removals or resets for an ideal layout. The puzzle's score is judged by the $\text{Score} = (\text{FW} + 0.5 \cdot \text{LL}) \times \text{FR} \times  \text{LR}$
where, $\text{FW}$ denotes the total words inserted, $\text{LL}$ is the count of crossing letters, $\text{FR}$ represents the proportion of filled space, and $\text{LR}$ the density of intersecting letters. End criteria include reaching necessary puzzle complexity, obtaining a given space ratio, limiting grid adjustments, or maximizing allocated time, all ensuring a targeted and time-bound puzzle output. It emphasizes adaptability in honoring preferred answers to align with specific educational aims or content themes.

\section{Experiments} \label{sec:Experiments}
In this section, we describe the experimental evaluation of two methods for generating Turkish crossword clues from keywords and their text categories. We detail the datasets used for training, the experimental setup for each method, and the methodologies implemented. Finally, we showcase a Turkish crossword created using our methodology as an example of its application.
\paragraph{Generating the Crossword Clues From Given Answer:}
In our study on automated crossword clue generation, described in Section \ref{sec:dataset}, we employed a \textit{TAC} dataset sub-set consisting of 60,000 pairs provided by experts. We fine-tuned \textit{GPT3.5-Turbo} on a chosen subset with a batch size of 16 and a learning rate of 0.01 for three epochs, targeting Turkish crossword clues from keywords. We tested the model's performance using 2,135 academic keywords to assess its capability to produce pertinent and creative clues.
The evaluation of generated crossword clues was conducted by two native Turkish language speakers who served as human judges.
This process revealed that 51.8\% of the clues achieved the set standards for acceptability, highlighting the model's capability to create contextually relevant and innovative clues. This result emphasizes the model's potential for further improvements and its applicability in educational tools to enhance learning in diverse subjects.  
\paragraph{Generating Clues From Given Text Answer Category:}
To analyze specific queries from the text, we used a dataset depicted in Section \ref{sec:data2}, refining it down to 8670 unique, clue-paired texts. We designated 8000 of these for training and 670 for testing. We fine-tuned \textit{GPT3.5-Turbo}, \textit{Llama-2-7b-chat-hf} and \textit{Llama-2-13b-chat-hf} models; \textit{GPT3.5-Turbo} with a batch size of 16, a learning rate of 0.001 over three epochs, \textit{Llama-2-7b-chat-hf} and \textit{Llama-2-13b-chat-hf} using Parameter Efficient Fine-Tuning (PEFT) with r=16, alpha=32, and a learning rate of 0.0001 across three epochs for both models. Model performances for generating clues from the test set were assessed via ROGUE-1, ROGUE-2, and ROGUE-L F1 score metrics, compared to the generated clues from \textit{GPT4-Turbo}, with detailed results in Table 
\ref{tab:llms_results}.

\begin{table}[ht!]
     \centering
     \small
     \begin{tabular}{cccccc}
      \hline
    \textbf{model type}&    \textbf{model name} & \textbf{\# params} & \textbf{ROUGE-1}& \textbf{ROUGE-2} & \textbf{ROUGE-L} \\ \hline
     &\textsc{Llama2-chat}  & 7B & -- & -- & --\\
   Base LLMs &\textsc{Llama2-chat}  & 13B & -- & -- & --\\
    &\textsc{GPT3.5 Turbo}  & - & 24.57 & 8.48 & 16.61\\
     \hline
   
    &\textsc{Llama2-chat}  & 7B & 22.50 & 5.00 & 14.80\\
   Finetuned LLMs
    &\textsc{Llama2-chat}  & 13B & 25.66 & 6.45 & 17.24\\
    
    &\textsc{GPT3.5 Turbo}  & - & \textbf{39.96} & \textbf{18.08} & \textbf{23.48}\\
    \hline
     \end{tabular}
     \caption{Performance of LLMs with and without fine-tuning.}
     \label{tab:llms_results}
 \end{table}

\begin{figure}[ht!]
\centering
  \includegraphics[width=0.9\linewidth]{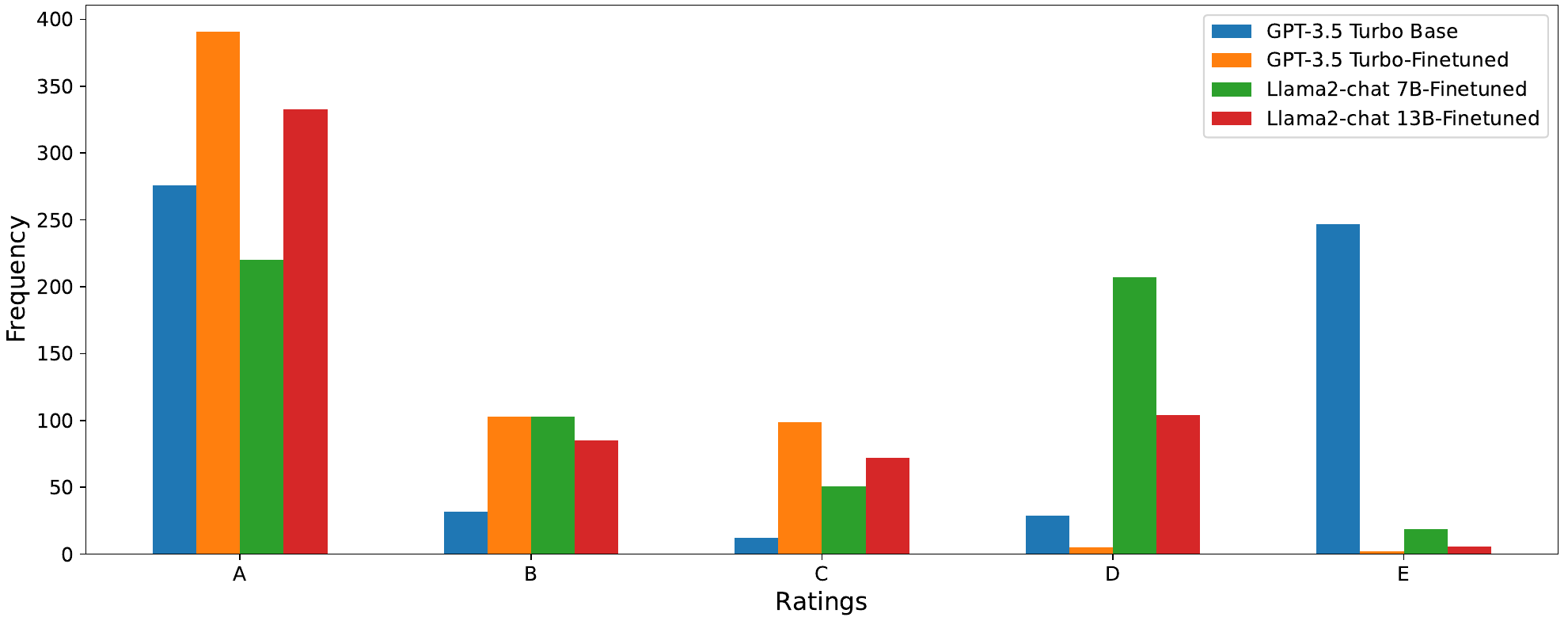}
  \caption{Comparison of model ratings.}
  \label{fig:comparison_of_models}
\end{figure}
\begin{figure}[ht!]
    \centering
       \includegraphics[width=0.9\textwidth]{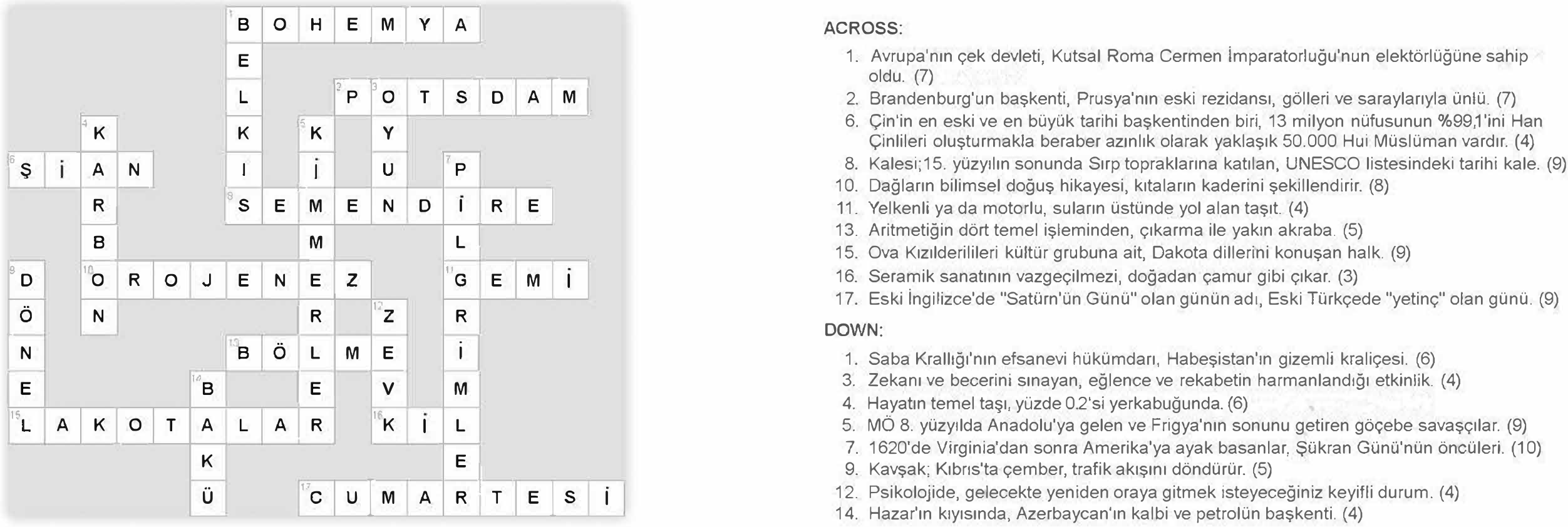}\\
    \caption{Crossword crafted using the proposed approach.}
    \label{fig:crossword}
\end{figure}

In our study, we evaluated \textit{GPT3.5-Turbo}, \textit{Llama-2-7b-chat-hf}, and \textit{Llama-2-13b-chat-hf} models on their ability to generate Turkish crossword clues using the \textit{T4TAC} dataset. The base models of \textit{Llama-2-7b-chat-hf} and \textit{Llama-2-13b-chat-hf} initially showed limited capability in processing Turkish. To analyze the impact of fine-tuning on model performance, we conducted a detailed human evaluation using 200 texts from both the base and fine-tuned versions, examining three clues per text, totaling 600 examples. Native Turkish speakers carried out the evaluations to ensure accurate assessment in terms of linguistic and cultural relevance. The results, depicted in Figure \ref{fig:comparison_of_models}, indicate a clear enhancement in performance post-fine-tuning with the \textit{T4TAC} dataset. Both \textit{Llama} models initially struggled with Turkish crossword clues but improved markedly after fine-tuning, as did the \textit{GPT3.5-Turbo} model.\\
We examined two crossword clue creation methods: one with set answers and another categorizing text-generated clues. Both support custom clues for educational goals, allowing educators to choose optimal clues for crossword construction (see Section \ref{sec:schema}). A sample Turkish crossword crafted using this system is depicted in Figure \ref{fig:crossword}.

\section{Conclusion} \label{sec:conclusions}

In this study, we introduced the Turkish Educational Generator, a novel tool powered by LLMs, to create dynamic crosswords in Turkish for educational purposes. This system allows educators to easily generate subject-specific crosswords, boosting student engagement and learning retention. Additionally, we enriched Turkish language datasets by providing two comprehensive datasets: one from expert-crafted answer-clue pairs and another from generated clues with corresponding text, answers, and categories, both vital for educational system development and research. Future plans include expanding our tool to more languages and enhancing clue-generation techniques with advanced LLMs, pushing the boundaries of educational technology.

\end{document}